\documentclass[a4paper]{article}
\usepackage{url}
\usepackage{INTERSPEECH2022}

\title{Annotated Speech Corpus for Low Resource Indian Languages:\\ Awadhi, Bhojpuri, Braj and Magahi}
\name{Ritesh Kumar$^1$, Siddharth Singh$^1$, Shyam Ratan$^1$, Mohit Raj$^1$, Sonal Sinha$^1$, Bornini Lahiri$^2$, Vivek Seshadri$^3$, Kalika Bali$^3$, Atul Kr. Ojha$^4$$^,$$^5$}
\address{
  $^1$Dr. Bhimrao Ambedkar University, Agra\\
  $^2$Indian Institute of Technology-Kharagpur\\
  $^3$Microsoft Research India, Bangalore\\
  $^4$Panlingua Language Processing LLP, New Delhi\\
  $^5$National University of Ireland, Galway}
\email{riteshkr.kmi@gmail.com}

\begin{document}

\maketitle
\begin{abstract}
In this paper we discuss an in-progress work on the development of a speech corpus for four low-resource Indo-Aryan languages - Awadhi, Bhojpuri, Braj and Magahi using the field methods of linguistic data collection. The total size of the corpus currently stands at approximately 18 hours (approx. 4-5 hours each language) and it is transcribed and annotated with grammatical information such as part-of-speech tags, morphological features and Universal dependency relationships. We discuss our methodology for data collection in these languages, most of which was done in the middle of the COVID - 19 pandemic, with one of the aims being to generate some additional income for low-income groups speaking these languages. In the paper, we also discuss the results of the baseline experiments for automatic speech recognition system in these languages.

\end{abstract}


\section{Introduction}
Development of reliable speech technology for several low-resource languages of India has always been a challenge. Although many automatic speech recognition systems (ASR) \cite{Malik2021} have been built for some major languages of India, still there is lack of appropriately transcribed speech corpus and models for several other languages \cite{Basu2021, model-survey}. To overcome the lack of these low resources of data, many initiatives have been taken by several projects and teams in recent times. For example, Interspeech 2018 Low Resources ASR challenge \cite{inproceedings} released speech data of phrasal and conversational speech, with transliteration of 50 hours each in Gujarati, Tamil and Telugu. 

More recently \cite{Basu2021} discusses the development of a speech corpus for 16 low-resource languages of Eastern and North-Eastern India including those for languages like Adi, Angami, Ao,
Hrangkhawl, Khasi, Lotha, Mizo, Nagamese, Sumi and others. The speech data, collected from 240 native speakers, totals around 2.17 to 6.67 hrs of each language. The data was used for developing speaker and language identification systems for these languages are assessed using different feature-classifier combinations of 10s duration. 

There are some other speech corpora prepared by group of researchers and some organisations under different projects: 
\begin{itemize}
    \item Hindi speech database consists of 500 spoken sentences by 50 speakers developed for speech recognition system, study of acoustic - phonetic correlates of Hindi and speaker recognition. Further, these sentences are divided into two sets: Set one has 2 sentences covering most of the Hindi phonemes and set two has 8 sentences covering most of the phonemic contexts. This data was manually segmented and labeled with sub-phonetic units and adequately exhaustive to get acoustic, phonetic, inter-speaker and intra-speaker variabilities in Hindi language \cite{samudravijaya00_icslp}.
    \item A standard speech corpora is developed with the purpose of developing ASR systems for three Indian languages: Hindi, Indian English and Bengali. This corpora was recorded by 1,500 male and female speakers of different age groups for each language on a online recording system and transcribed by semi-automatic tools \cite{basu:2019}. 
    \item Bangali speech corpus is developed with the help of recording Anandabazar text corpora read by 40 female and 70 male speakers. The text corpus has 7,500 unique sentences, 19,640 unique words. 26 hours of speech corpus is developed by two age groups: younger group i.e. 20 to 40 years and older group i.e. 60 to 80 years. This speech corpus is labeled at phone and triphone level \cite{das2011bengali}.
    \item Phonetic and Prosodically Rich Transcribed (PPRT) Speech corpus in Bengali and Odiya languages is developed. It consists of five hours of conversation speech, ten hours of read speech and five hours of extempore speech \cite{sunil:2013}.
    \item IITKGP-MLILSC (Indian Institute of Technology Kharagpur - Multilingual Indian Language Speech Corpus) speech database consists of data from 27 Indian languages: Arunachali, Assamese, Bengali, Bhojpuri, Chattisgarhi, Dogri, Gojri, Gujrati, Hindi, Indian English, Kannada, Kashmiri, Konkani, Manipuri, Mizo, Malayalam, Marathi, Nagamese, Nepali, Oriya, Punjabi, Rajasthani, Sanskrit, Sindhi, Tamil, Telugu and Urdu. This corpus is recorded from television and radio channels: news bulletins, live shows, interviews, and talk shows. 5-10 minutes of data is collected from each source. Broadly, minimum of 1 hour speech data for each language is collected \cite{maity:2012}.
    \item Speech data for Bangla is collected for the development of an Interactive Voice Response (IVR) based Commodity Price Retrieval System in Bangla for farmers \cite{basu:2013}. 
    \item Speech corpora for low-resourced languages of North-East India is designed for Assamese, Bengali and Nepali. 1,000 sentences of Assamese language from novels, story books and proverbs were read and recorded by 27 native speakers on telephone channel with the help of interactive voice response system \cite{deka:2018}. 
    \item Database for Mizo tones is collected from five Mizo speakers from Aizwal area of Mizoram. 
    In total 4,384 syllables are collected form 338 three syllable/word sentences presented by each speaker. All 4,384 syllables are annotated by native speaker for tones in Mizo by visual and listening examination \cite{Sarma2015DetectionOM}.
    \item ALS-DB (Arunachali Language Speech Database) is a multilingual and multichannel speech database of languages from Arunachal Pradesh. 
    100 female and 100 males of 20-50 years age group participated in recording the speech data. They read story from the school book of 4-5 minutes in each language. Every speaker recorded for three languages: Hindi, English and one of the four local languages: Apatani, Adi, Galo, and Nyishi \cite{sarmah:2013}.
    \item Marathi speech database is a continuous speech database developed for Marathi language. It is collected from 34 districts of Maharashtra from 1500 literate speakers. These speakers are selected on the basis of gender, age group, educational qualification and mother tongue for speech data collection. Speakers prepared this data by reading sets of sentences. There are some features extracted from this database are phonemic richness, average length of words and sentences, gender distribution of speakers and amount of non speech sounds \cite{godambe:2013}.
    \item Hindi and Marathi corpus is prepared under the project of Government of India ``Speech-based Access for Agricultural Commodity Prices in Six Indian Languages". In this corpus Hindi and Marathi has 1 hour and 8 hours of data. This corpus is collected with real time environment like quiet indoor home environment as well as outdoor environment with vehicle and machine noise. It mainly consists of names of local districts and agricultural commodities \cite{MOHAN2014167}.
    
\end{itemize}

Although Automatic Speech Recognition (ASR) has done remarkable progress in last few years because of the availability of speech corpora of various kinds for ``official" and relatively richer languages, it is still an extremely challenging tasks for non-scheduled, extremely low-resourced languages, generally spoken by people with low income because of the absence of any datasets for these languages. 
This situation is further exacerbated by the fact that there is a rather minimal government or industrial funding or support for most of these languages because of various reasons. This has led to a situation where millions of people do not have access to speech and language technologies in their own language and may not have access in the next several years.

In this paper, we discuss an in-progress work of the development of a speech dataset in four Indo-Aryan languages - Awadhi, Bhojpuri, Braj and Magahi - using field methods of linguistic data collection, remodelled as limited crowdsourcing-like micro-tasks. This method proved to be advantageous in three ways (see, however, Section \ref{sec:challenges} for challenges and issues with this) -

\begin{itemize}
    \item It allowed us to collect the data remotely during the COVID-19 pandemic.
    \item It provided the (mostly low-income) speakers an opportunity to make an earning during the extremely difficult period of the pandemic. 
    \item It led to the development of speech datasets for hitherto largely ignored langauges.
\end{itemize}
In the paper, we also discuss some of our baseline experiments with the dataset that we have completed till now and demonstrate how even a small dataset like ours could give significant improvement in ASR for these languages over a zero-shot system based on a high-resource closely-related language like Hindi.

\section{Our Dataset}

\begin{table*}[!h]
    \centering
    \begin{tabular}{c|c|c|c|c}
    \hline
    \hline
    \textbf {Details} & \textbf{Awadhi} & \textbf{Bhojpuri} & \textbf{Braj} & \textbf{Magahi} \\
    \hline
    \hline
    Region & Pratapgarh, Uttar Pradesh (UP) East & Patna, Sasaram, Betiah, Banaras, Ballia & Agra, UP West & Patna District\\ 
    \hline
    Gender & 5 Female, 5 Male & 7 Female, 3 Male  & 5 Female, 5 Male & 5 Female, 5 Male\\
    \hline
    Age (yrs) & 18 - 35 & 24 - 75 & 18 - 30  & 18 - 35 \\
    \hline
    Lingual & Multi & Mono-Multi & Multi & Multi \\
    \hline
    \end{tabular}
    \caption{Speaker Details}
    \label{tab:details}
    \vspace{-6mm}
\end{table*}


\subsection{Speaker Selection}

We have collected speech data for four languages - Awadhi, Bhojpuri, Braj, and Magahi given in Table \ref{tab:dataset}. We have recorded speech data via Karya app - a mobile-based crowdsourcing tool, especially aimed at generating income for low-income groups by providing them opportunities to work on for crowd-based tasks \cite{abraham2020crowdsourcing}. The process of sampling the speakers took into consideration the following criteria.\par
\begin{itemize}
    \item Age – The age of the language expert should be more than eighteen years.
    \item Gender – The number of male and female language experts should be roughly equal for each language.
    \item Region – They should have spent most of their time in the same region from where the data was being collected - this was needed to avoid excessive influence of Hindi on these languages.
    \item Language – The speakers use their native language in the home domain - again this was needed since all of the languages under study have witnessed a shift to Hindi in the major urban centres, especially among the educated population .
\end{itemize}

We have selected a total of 40 language experts, ten from each language, by considering the above-mentioned linguistic eligibility criteria (see Table \ref{tab:details}). In order to collect clean and noise-free data the speakers were asked to record speech in the absence of any noise like the sound of the fan, birds chirping, dog barking, family chattering etc. 

\subsection{Data Collection via Field Methods}
We recorded the speech using the methods used by field linguists for linguistic data collection. The rationale for using the field methods for data collection instead of the usual method of recording read or narrated speech using some random sets of texts from web sources or other texts is explained here.

The questionnaires designed by field linguists for linguistic data collection are generally prepared (and perfected over years of fieldwork in hundreds of languages across the globe) with lot of care and attention such that the data collected from using these questionnaires could be utilised for grammatical description of specific phenomena and, possibly, of the language as a whole. This is not generally possible with any random text.
Since the languages under the current study are not only under-resourced but are underrepresented, minoritised and devalued (by being mistakenly referred to as ``dialects" of Hindi despite very robust linguistic studies demonstrating that it is both historically and structurally erroneous to make this claim), the speakers themselves hesitate in using the language in public, especially when they are being recorded. In such a situation, it could be assumed that it would be extremely difficult to collect data from these languages again - it could also be gauged from the fact that there is hardly any resource available for these languages prior to our work. As such we wanted our data to be useful not only for just one task - Automatic Speech Recognition - but to be more generally useful for the larger community of NLP practitioners, linguists, speakers and other stakeholders.

As a common practice in field-based data elicitation methods, our initial attempt at data collection involved two phases - translation phase and narration phase. In the translation phase, we provided 369 sentences in Hindi and asked the speakers to translate those into the respective languages viz. Awadhi, Bhojpuri, Braj, and Magahi. These sentences are Hindi translation of the questionnaire prepared by \cite{ju-quest} for collecting data for language documentation and description in Indian languages. These sentences belong to the domain of daily routine life situations like domestic work, food, cooking, etc. More importantly, these sentences have been designed to elicit patterns of a large number of grammatical phenomena such as case, classifiers, reflexives and reciprocals, tense, mood and aspect, ECV, reduplication and others. Thus it allowed us to not only get data for describing the different phenomena in the language but we also get representation of various kinds of sentences in the dataset.

In the narration phase, we prepared 39 questions related to the three main lifecycle events in our culture - birth, marriage, and death. These questions were prompts asking the speakers to talk about their rituals and tradition related to these events. This yielded a more naturalistic narrative data (in comparison to the translated sentences elicited in the first phase) and also allowed us to collect those kinds of sentences that were not possible to elicit via the translations.

Both the questionnaires (and all the future questionnaires) used for data collection in the project will be made publicly available, along with the dataset (if they are not already available).

\subsection{Data Transcription and Annotation}
After the completion of speech recording, we have sliced the speech signal based on sentence completion and then transcribed it into Devanagari script and exported the data in TextGrid format. The transcribed data is then further exported to the CONLL-U format and annotated with part-of-speech labels, morphological features and Universal Dependencies relation. The statistics of the transcribed and annotated dataset is give in Table \ref{tab:datasetcounts}.


\begin{table*}[!h]
    \centering
    \begin{tabular}{c|c|c|c|c}
    \hline
    \hline
    \textbf{Language} & \textbf{Translation Sentences} & \textbf{Translation Tokens} & \textbf{Narration Sentences} & \textbf{Narration Tokens} \\
    \hline
    \hline
    \textbf{Awadhi} & 2,320 & 15,692 & 620 & 16,601\\ 
    \textbf{Bhojpuri} & 2,466 & 16,228 & 482 & 7,705\\
   \textbf{Braj} & 1,057 & 6,961 & 1,298 & 21,216 \\
    \textbf{Magahi} & 2,311 & 15,797 & 630 & 10,563 \\
    \hline
    \textbf{Total} & \textbf{8,154} & \textbf{54,678} & \textbf{3,030} & \textbf{56,085} \\
    \hline
    \end{tabular}
    \caption{The Speech Dataset Counts}
    \label{tab:datasetcounts}
    \vspace{-4mm}
\end{table*}

\section{How does this dataset help?}

As mentioned earlier, since these languages are often mistaken to be the ``varieties" of Hindi, it is generally assumed that the systems built for Hindi should also work fine for these languages. This is expected to be true even if these languages are considered closely-related to Hindi but not its variety. As such we conducted some experiments to understand if this is indeed the case and how well Hindi ASR models perform for these languages. We then also compared the performance of these Hindi models with some baselines that have been developed for these languages - either from scratch or using transfer learning. We used the two commercially available Hindi ASR systems for transcribing the dataset in all of the 4 languages and calculating their WER on our test set -

\begin{itemize}
    \item Speech-to-text API by Google Cloud\footnote{\url{https://cloud.google.com/apis/docs/overview}}
    \item Speech-to-text API by Azure Cognitive Services (Microsoft)\footnote{\url{https://azure.microsoft.com/en-in/}}
\end{itemize}

In the second step, we augmented these models with the language-specific vocabulary and language model to see if this helps or not. Finally in the third step, we trained models for these languages using two approaches -

\begin{itemize}
    \item Training from scratch: We used Kaldi recipes \cite{kaldi:2011} provided by \cite{Diwan_2021}\footnote{the script is available here: \url{https://github.com/navana-tech/baseline_recipe_is21s_indic_asr_challenge/blob/master/is21-subtask1-kaldi/s5/run.sh}} to train the ASR models for these languages from scratch. We experimented with four different models - monophone model (mono), triphone model (tri1), triphones with Delta feature augmentation (tri2b) and triphones with both delta feature augmentation and speaker normalisation (tri3b).

    \item Transfer learning: We fine-tuned wav2vec-large-xlsr-53 \cite{conneau:2020} model using the complete, multilingual dataset and evaluated the performance of the model for the whole test set.
\end{itemize}

For both the approaches. we used two kinds of setups for training and evaluation. In the first setup, we trained monolingual models of each of the languages and calculated average WER of these. In the second setup, we trained a multilingual model and evaluated all the languages together.  This was also meant to test if a single multilingual model gives a performance improvement over multiple monolingual models even with this small dataset or not.
We report the WER for each of these models for each language in Table \ref{tab:dwer}.

As is evident, WER obtained on at least one of the Hindi models for all the languages is better than the models trained for these languages from scratch. However, the wav2vec2.0 model (transfer learning) for each of the language is almost half of those for the models trained from scratch or the Hindi models. Moreover, in all cases, in both the models trained from scratch as well as those trained using transfer learning, multilingual models outperform the multiple monolingual models. These results are on expected lines.

\begin{table*}[!h]
    \centering
    \begin{tabular}{c|c|c|c|c|c|c}
    \hline
    \hline
    \textbf{Models} & \textbf{Awadhi} & \textbf{Bhojpuri}  & \textbf{Braj} & \textbf{Magahi} & \textbf{Avg. WER} &  \textbf{Multilingual} \\
    \hline
    \hline
    \textbf{Azure Hindi} & 83.28 & 76.84 & 91.08 & 83.38 & 83.64  & -\\ 
    \textbf{Google Hindi} & \textbf{79.93} & 67.06 & 82.89 & 77.53 & 76.85 & -\\
  
 
    \textbf{mono} & 86.37 & 82.76 & 93.25 & 88.45 & 87.70 & 87.54\\
    \textbf{tri1} & 80.97 & 77.93 & 90.14 & 84.23 & 83.31 & 81.21\\
    \textbf{tri2b} & 82.16 & 79.76 & 90.38 & 82.53 & 83.70 & 81.17\\
    \textbf{tri3b} & 82.56 & 79.40 & 89.25 & 82.55 & 83.44 & 82.34\\
    \textbf{wav2vec 2.0} & - & \textbf{37.65}  & \textbf{56.78} & \textbf{38.03} & 
    \textbf{44.15} & \textbf{40.44} \\
    
    \hline
    \end{tabular}
    \caption{WER of the Baseline Models}
    \label{tab:dwer}
    \vspace{-5mm}
\end{table*}

\begin{table}[!h]
    \centering
    \begin{tabular}{c|c|c|c}
    \hline
    \hline
    \textbf{Language} & \textbf{Translation} & \textbf{Narration} & \textbf{Total} \\
    \hline
    \hline
    \textbf{Awadhi} & 01:52:29 & 02:54:01 & 04:46:30  \\ 
    \textbf{Bhojpuri} & 01:55:32 &  02:56:06 & 04:51:38\\
   \textbf{Braj}  & 02:14:59  & 02:20:01 & 04:35:00 \\
    \textbf{Magahi} & 02:27:27 & 01:20:16 & 03:47:43 \\
    \hline
    \textbf{Total} & \textbf{08:30:27} & \textbf{09:30:24} & \textbf{18:00:51} \\
    \hline
    \end{tabular}
    \caption{The Speech Dataset}
    \label{tab:dataset}
    \vspace{-9mm}
\end{table}

\section{Challenges}
\label{sec:challenges}
The human speech and its variations due to different dialects and accents, social factors like gender, age and speed of utterance create challenges in building ASR systems. Therefore it is generally recommended to keep the datasets as balances as possible in these different ways.
In our case, in addition to finding this balance among these factors, use of the `Karya' app, especially during the pandemic, proved to be tricky and challenging. It was not easy for those speakers who were not acquainted with smartphones to understand the workings of even a relatively simple app like Karya. This resulted in occasional incomplete and corrupted data. Moreover, the recording was done by the users themselves (without any significant guidance or training), thus there were some lack of expertise as how to use the mobile speaker adequately for recording tasks like these. Some speakers' voice were too low or murmuring and breaking and not usable; in others silences were too long before and after the actual recordings. Some of the recordings also contained noises of birds, vehicles, fans, or voices of some other people momentarily. 
These challenges had become more acute because of the inability of the researchers to actually visit the field, train the speakers in recording and then ask them to do the recordings. Unlike the earlier data collection efforts involving Karya, that was done via actual field visit, all the instructions had to be passed through the mobile phone in our case because of the restrictions imposed by the COVID-19 pandemic; thus there was no possibility of demonstrating the usage of the app. This has affected both the quality and the quantity of the dataset that have been collected till now. However, despite this, the app did enable us to collect and transfer the data even in the middle of the pandemic, when there was no other means of collecting data. On the other hand, the possibility of extra income that the task provided for the low-income groups in a very difficult time, proved to be an additional incentive for the completion of the task.

\section{Ethical and Societal Implications}


Since the data for this research is collected directly from the speakers, it involved the usual ethical considerations for working with speech communities for linguistic data collection. It included explaining the purpose of the data collection, how the data will be used, speakers' intellectual property rights over the data contributed by them and all its derivatives and finally getting their informed consent for using the dataset for research. In addition to this, more specific societal implications of the research is discussed below -

    \textbf{Short-term good of the speakers' community}: As we mentioned earlier the data was collected in the middle of the pandemic and it provided the speakers, who were all selected from the low-income groups \footnote{Since Karya is designed to provide income to low-income groups, it was a conscious decision on our part. However, along with this, since these languages are now generally not preferred to be spoken by the ``elite", urban, educated population, it also proved to be a pre-requisite to get the data that we were aiming to collect.}, an opportunity for some additional income in very trying and difficult times. Although this was nowhere substantial, we would like to think that it did help the speakers in some minimal way.
    
    \textbf{Possible enhancement of Language Prestige}: Since all of the languages under study are relegated to the status of `illiterate', `uncouth' or `rural' varieties of Hindi, the speakers do not generally feel very comfortable owning up to the language or speaking it for ``official" purposes like ours. However, our interest and insistence on this very ``version" of the language rubbed a little on the speakers as well and led to a possible enhancement of language prestige, with many speakers admitting that they never thought that it could be of any use for highly educated people like University students and professors. Moreover, it was also surprising and encouraging that they could earn some money by knowing and speaking their own language (as opposed to English or Hindi).
    
    \textbf{The Dataset and its application}: Finally, of course, as the dataset is further augmented, it could lead to the development of usable speech technologies for millions of people such that they could access the technologies and content in their own language, without the need to switch to Hindi. Moreover, since a large population of speakers of these languages may not be able to access the writing-based content and technologies, working speech-based technologies might prove to be an essential tool for them.

\section{Summing Up}
In this paper, we have discussed an in-progress work on the development of an annotated speech corpus and baseline ASR systems for 4 extremely low-resource Indian languages, spoken largely by low-income groups. We have demonstrated how even a small dataset like ours could prove to be effective in building a reasonably better system for these languages. We have also argued that our research has the potential to have a positive social impact at multiple levels.


\bibliographystyle{IEEEtran}

\bibliography{mybib}


\end{document}